\documentclass[sigconf, nonacm]{acmart}

\usepackage{graphics} % for pdf, bitmapped graphics files
\usepackage{amsmath} % assumes amsmath package installed
\usepackage{xcolor}
\usepackage{url}
\usepackage{tikz}
\usepackage{tikz-network}
\usepackage{booktabs}
\usepackage[tableposition=top]{caption}

\AtBeginDocument{%
  \providecommand\BibTeX{{%
    \normalfont B\kern-0.5em{\scshape i\kern-0.25em b}\kern-0.8em\TeX}}}

\acmYear{2023}
\acmDOI{}

\acmConference[]{2023}{}
\acmPrice{}
\acmISBN{}
\setcopyright{none}

\newcommand{\beginsupplement}{
  \setcounter{table}{1}  
  \renewcommand{\thetable}{S\arabic{table}} 
  \setcounter{figure}{1} 
  \renewcommand{\thefigure}{S\arabic{figure}}
  \setcounter{section}{1} 
  \renewcommand{\thefigure}{S\arabic{figure}}
}

\begin{document}

% \title{Neuromorphic Flight Control of a Miniature Drone using Spiking Neural Networks}
\title{Neuromorphic Control using Input-Weighted Threshold Adaptation}
% \title[A tiny drone with a spiking neural regulator]{A tiny drone with a spiking neural regulator}

% "Neuromorphic Flight Control of a Miniature Drone using Spiking Neural Networks"

\author{Stein Stroobants}
\orcid{0000-0001-5733-1677}
\email{s.stroobants@tudelft.nl}
\affiliation{%
  \institution{Delft University of Technology}
  \streetaddress{Kluyverweg 1}
  \city{Delft}
  \country{The Netherlands}
  \postcode{2629HS}
}

\author{Christophe De Wagter}
\orcid{0000-0002-6795-8454}
\affiliation{%
    \institution{Delft University of Technology}
  \streetaddress{Kluyverweg 1}
  \city{Delft}
  \country{The Netherlands}
  \postcode{2629HS}}

\author{Guido C.H.E. de Croon}
\orcid{0000-0001-8265-1496}
\affiliation{%
  \institution{Delft University of Technology}
  \streetaddress{Kluyverweg 1}
  \city{Delft}
  \country{The Netherlands}
  \postcode{2629HS}
}

\renewcommand{\shortauthors}{Stroobants, et al.}

\begin{abstract}
% The promising features of neuromorphic processing, such as high energy efficiency and response rates, encourages the development of neuromorphic control algorithms for tiny robotics. 
% However, low-level controllers on these systems are mostly unconsidered and overlooked, and their contribution to the stability of dynamic systems is undervalued. 
% Additionally, even the simple task of integrating errors over time has shown to be difficult in these networks. 
Neuromorphic processing promises high energy efficiency and rapid response rates, making it an ideal candidate for achieving autonomous flight of resource-constrained robots. It will be especially beneficial for complex neural networks as are involved in high-level visual perception.
% Especially high-level complex algorithms using for instance visual data could benefit. 
% However, for these algorithms, it is extremely important to also look at the low-level controllers that are critical to ensure the stability of dynamic systems.
However, fully neuromorphic solutions will also need to tackle low-level control tasks. 
% Current state-of-the-art focuses on the big picture but often overlooks the low-level controllers that are critical to ensuring the stability of dynamic systems. 
Remarkably, it is currently still challenging to replicate even basic low-level controllers such as proportional-integral-derivative (PID) controllers. Specifically, it is difficult to incorporate the integral and derivative parts.
% Without reliable solutions for the integral and derivative pathways, the full potential of sensor-to-actuator neuromorphic control can be limited. 
% This gap in understanding can limit the full potential of sensor-to-actuator neuromorphic control, which requires reliable solutions for these overlooked components.
% CDW: overlooked is erg negatief:
% liever: integral and derivative are still challenging 
%
% To allow for full sensor-to-actuator neuromorphic control, reliable solutions for these need to be realized.
% This work proposes a highly modular neuromorphic controller with an architecture that promotes proportional, integral and derivative pathways during learning. 
% Especially, an input threshold adaptation mechanism was proposed for the integral pathway. 
To address this problem, we propose a neuromorphic controller that incorporates proportional, integral, and derivative pathways during learning. 
Our approach includes a novel input threshold adaptation mechanism for the integral pathway. This Input-Weighted Threshold Adaptation (IWTA) introduces an additional weight per synaptic connection, which is used to adapt the threshold of the post-synaptic neuron.
We tackle the derivative term by employing neurons with different time constants.
% To exhibit the value of this network, it was implemented on a microcontroller connected to the open-source tiny Crazyflie quadrotor. 
% The network here replaced the inner-most rate controller and its stability was demonstrated during manual flight. 
% Our results show that these bio-inspired algorithms can be used to control highly dynamic systems on a very low-level under the influence of disturbances. 
We first analyze the performance and limits of the proposed mechanisms and then put our controller to the test by implementing it on a microcontroller connected to the open-source %here open-source is an adjective to quadcopter -> hyphen
tiny Crazyflie quadrotor, replacing the innermost rate controller. We demonstrate the stability of our bio-inspired algorithm with flights in the presence of disturbances. 
The current work represents a substantial step towards controlling highly dynamic systems with neuromorphic algorithms, thus advancing neuromorphic processing and robotics. 
In addition, integration is an important part of any temporal task, so the proposed Input-Weighted Threshold Adaptation (IWTA) mechanism may have implications well beyond control tasks.
% These promising results highlight the potential of our approach for controlling highly dynamic systems at a low level. 
% By shining a spotlight on the importance of low-level controllers, our work opens up new possibilities for advancing neuromorphic processing and robotics.
\end{abstract}


%   <concept>
%       <concept_id>10010147.10010178.10010213.10010214</concept_id>
%       <concept_desc>Computing methodologies~Computational control theory</concept_desc>
%       <concept_significance>300</concept_significance>
%       </concept>
%  </ccs2012>

% \ccsdesc[300]{Computing methodologies~Computational control theory}


%  \centering
%   \includegraphics[width=0.3\textwidth]{samples/images/neuromav.png}
%   \caption{Quadrotor used in this research equipped\\ with a neuromorphic processor to perform altitude control.}
%   \Description{}

\maketitle

\section{Introduction}

% Unmanned aerial vehicles (UAVs), also known as drones, have become increasingly popular over the years, thanks to their versatile applications in various fields. 
% One type of UAV that has gained significant attention in recent times is the quadrotor, a four-rotor aerial underactuated vehicle capable of maneuvering in tight spaces and performing complex tasks [REF]. 
% The use of quadrotors has extended to areas such as search and rescue, surveillance, agriculture, and delivery services, among others.
% However, limitations in size, weight and power, hamper the application of novel processing methods, such as complex neural networks for visual perception.
% Besides, while conventional control techniques have been successful in controlling quadrotors, they are still no match to control systems found in nature [REF]. 
% One fascinating example of such a biological control system is the haltere reflexes of the \textit{drosophila melanogaster}, which in combination with the visual system enables the fly to maintain stable flight in turbulent conditions \cite{dickinson1999haltere}. 
% Neuromorphic computing, an approach that seeks to emulate the behavior of biological neural networks in computing systems, has shown great promise in enhancing the performance of robotic control systems [REF]. 

Autonomous drones are envisaged for a wide range of applications. Many of these applications require high levels of autonomy. Currently, this is out of reach for many drones, since they are very limited in terms of size, weight, and processing~\cite{sekander2018multi}. Deep neural networks require heavy, power-hungry processors. That is why there is a surge of interest in bio-inspired, neuromorphic processing, which carries the promise of low-latency, energy-efficient processing of deep neural networks~\cite{schuman2017survey}. 
Although there is a lot of focus on complex neural networks for high-level visual perception~\cite{giusti2015machine}, a fully neuromorphic solution will need to encompass low-level control as well~\cite{abadia2021cerebellar}.
Remarkably, it is currently still highly challenging to replicate even simple low-level controllers such as PID controllers with spiking neural networks. 
An example of such a low-level controller is the fascinatingly elegant biological system that affects the haltere reflexes of the \textit{drosophila melanogaster}~\cite{dickinson1999haltere}.

Recently, an increasing amount of robotics research has been focused on developing spiking neural networks (SNNs) for control. % recently -> past tense
Specifically for controlling flying robots, examples include \citet{clawson2016spiking}, which uses reward-modulated synaptic plasticity to track a Linear-Quadratic Regulator (LQR) for a flapping wing drone. 
In \citet{qiu2020evolving}, a neuro-evolution strategy is utilized to learn a controller for a drone and is shown to outperform a PID in simulation.
Closer to our work are studies that mimic the behavior of conventional controllers with SNNs. % American English
Among those, the benefits of a spiking end-to-end control pipeline are especially clear in \citet{vitale2021event}. In this work, the rotation of a bi-rotor was controlled by combining a neuromorphic implementation of the Hough transform with a population-coded spiking implementation of the conventional PID controller.
They showed that due to the high update rates and asynchronous data flow from the event camera and neuromorphic chip, much faster responses could be obtained than with a conventional control setup.
The accuracy of this network scaled quadratically with the number of neurons, putting a limit on the resolution. 
\begin{figure}[b]
    \centering
    \includegraphics[width=\linewidth]{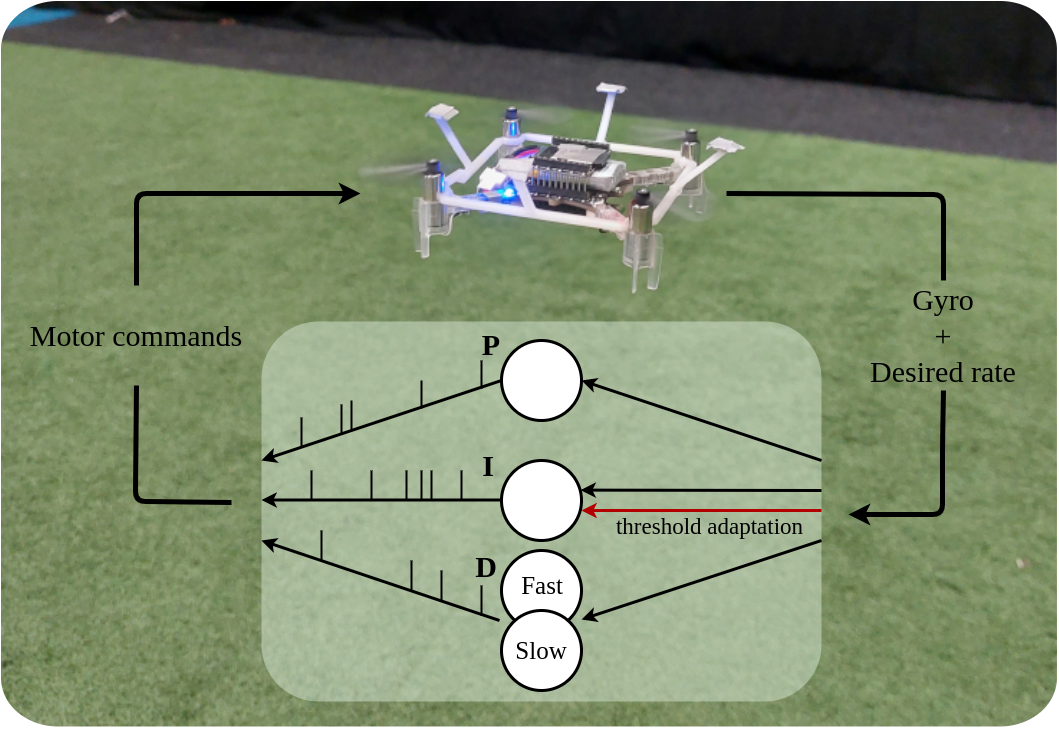}
    \caption{We propose a novel spiking neural network mechanism for realizing the integral term in a spiking PID controller and analyze the use of different time constants for the derivative term. For the integral term, we introduce Input-Weighted Threshold Adaptation, leading to a second weight per synapse. These mechanisms are demonstrated with onboard attitude rate control of a tiny Crazyflie drone.}
    \label{fig:crazyflie}
\end{figure}
In our previous work, the complexity of such a PID network was reduced to make it scale linearly with the number of neurons, and the network was used to control the altitude of a free-flying drone in a real-world test~\cite{stroobants2022design}. 
However, the integral and derivative paths in both these works showed clear limitations, imposed by the number of neurons used to represent the signals. 

Besides population coding, also rate coding was used to recreate conventional control. For example, in \citet{zaidel2021neuropid} the joints of a 6-DOF robotic arm were controlled by a rate-encoded PID controller, creating the integral pathway by having self-recurrent connections and the derivative by adopting different time constants.
Despite using this self-recurrency in the integral pathway, there still remained a steady-state error, showing the incapability of error integration over time.

In all these examples the seemingly simple tasks of calculating the integral and derivative over time have proven to be difficult for the current-based LIF model.
This may be due to the simplicity of the LIF neuron used in most robotics research. 
Popular for its simplicity, it fails to deal with the phasic-tonic response exhibited by biological neurons~\cite{levakova2019}.
Lastly, although all these works show clear steps toward a full end-to-end neuromorphic pipeline, none of them has demonstrated the ability to control the lowermost loop of a real, free-flying drone.

We present a neuromorphic controller that can more closely mimic PID controllers. We make the following contributions:
(1) We introduce Input-Weighted Threshold Adaptation (IWTA) to achieve more accurate integration. 
(2) We systematically study the capabilities and limitations of neurons with slow and fast time constants for obtaining the derivative term.
(3) We analyze the advantages and limitations of the introduced mechanisms.
(4) We demonstrate the novel neuromorphic controller with the onboard attitude rate control of a tiny ($\approx$30 gram) flying Crazyflie robot. 
Using only 320 neurons and a similar number of synapses per PID line, it is designed to take advantage of the unique characteristics of neuromorphic computing, such as high processing speed and fault tolerance, while maintaining the promise of low power consumption when implemented in specialized hardware.
The network is automatically trained to closely track the output of a conventional PID using backpropagation-through-time (BPTT), removing the necessity for manual tuning of network parameters. 

% We present the implementation of a neuromorphic controller for a tiny quadrotor. 
% Using only 320 neurons and a similar amount of synapses per PID line, it is designed to take advantage of the unique characteristics of neuromorphic computing, such as low power consumption, high processing speed, and fault tolerance. 
% It incorporates an encoding layer to transform floating point values to positive and negative spike trains, a proportional and derivative path and an integral group that uses the input signal to adapt its threshold. 
% In a simulated example, we show that this integral pathway
% We demonstrate the effectiveness of our approach through simulation and experimental results, highlighting its potential for real-world applications.

% By bridging the gap between biology and technology, our work opens up new avenues for the development of more sophisticated and efficient control systems for UAVs.

\section{Methods}
The entire SNN consists of multiple neuron groups, each resembling one of the separate parts of a conventional PID controller. 
All groups consist of an encoding layer, transforming floating point inputs to rate encoded spike-trains representing positive and negative values. 
These spikes propagate to neurons that respond proportionally to the input, to the accumulated signal over time or to the rate of change of the input. 
These independent systems are discussed in detail below.
% \begin{figure}[h!]
%     \centering
%     \includegraphics[width=0.7\linewidth]{images/network.png}
%     \caption{Overview of the network. The input value is encoded through the encoding blocks which immediately go to the corresponding proportional (P), integral (I) and derivative (D) blocks. {\color{red} Nog bedenken hoe ik dit duidelijk en compact kan laten zien (liefst de structuur duidelijk van alle connecties)}}
%     \label{fig:network}
% \end{figure}

\begin{figure}[h]
    \centering
    \includegraphics[width=0.9\linewidth]{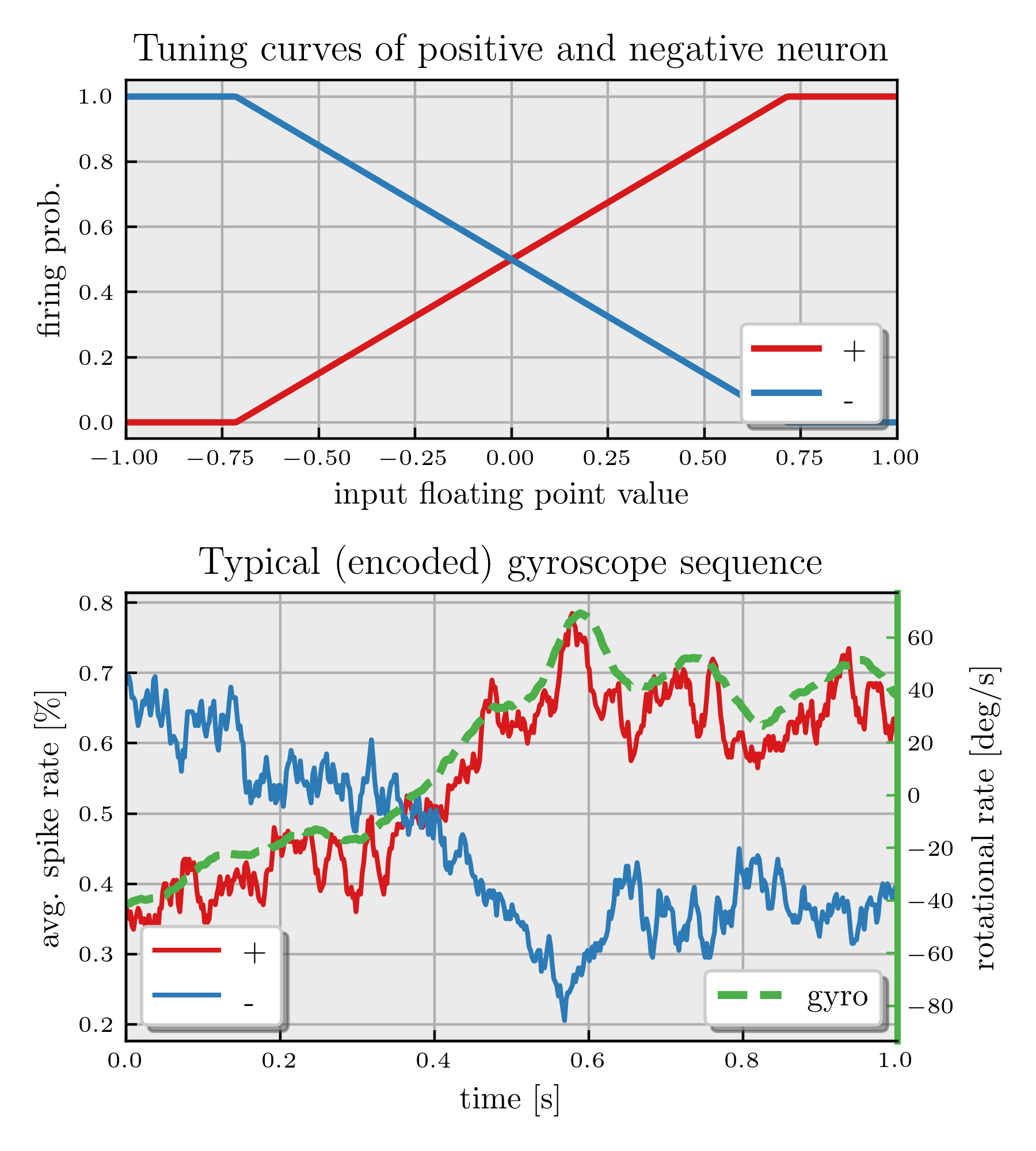}
    \caption{Overview of the encoding layer. The top graph shows the firing probability for a certain stimulus for positive and negative encoding neurons. The bottom graph depicts a typical sequence of encoded gyroscope measurements. The average spike rate of the encoded positive and negative spikes is shown in blue and red and the input sequence is shown in dashed green. }
    \label{fig:encoding}
\end{figure}

\subsection{Encoding - floating points to spikes}
For our network, a rate encoding scheme has been chosen that has separate channels for positive and negative values.
To ensure a certain spiking frequency at an input stimulus, encoding is done according to two (symmetrical) tuning curves. 
These tuning curves represent the spike probability of an encoding neuron to a given stimulus, and by comparing this to a random-generated number, either a spike (1) or not (0) is produced by the neuron. 
Although more complex functions can be chosen for these tuning curves, in this work a linear relation between spiking frequency and stimulus was chosen. 
This linear relation between input $i(t)$ and output spike probability $P(s(t))$ at time $t$ is dictated by the parameters $\alpha$ and $\beta$ as follows:
\begin{align}
    P(s(t)) &= \begin{cases}
        0 & r \beta i(t) + \alpha \leq 0 \\
        r \beta i(t) + \alpha & 0 \leq r \beta i(t) + \alpha \leq 1 \\
        1 & r \beta i(t) + \alpha \geq 1 \\
    \end{cases} \\
    r &= \begin{cases}
    1 & \text{positive neuron} \\
    -1 & \text{negative neuron}
\end{cases}
\end{align}
Where $r$ depends on whether the encoding neuron is positive or negative.
This is visualized in Figure \ref{fig:encoding} where the set of tuning curves used in this work and the resulting output spikes are shown for a typical input sequence measured with gyroscopes.

\subsection{Proportional - steering towards setpoint}
The output of the proportional layer should drive the system from its current state to a setpoint by responding linearly to the error. 
In this network, this is done by feeding the output of an encoding group, as described above, to two current-based Leaky Integrate-and-Fire (LIF) neurons, again each representing either positive or negative control commands (see Section S\unskip\ref{sm:cubalif} for the model used). 
The spikes from this layer are sent to an output leaky integrator, acting as an exponential filter to smooth the response. 
To ensure a balanced response, the synaptic weights in the network must be symmetrical for the positive and negative inputs.
By using more than one group, the stochastic effects induced by the encoding layer can be reduced and the accuracy increased. 
\begin{figure}[h]
    \centering
    \includegraphics[width=0.8\linewidth]{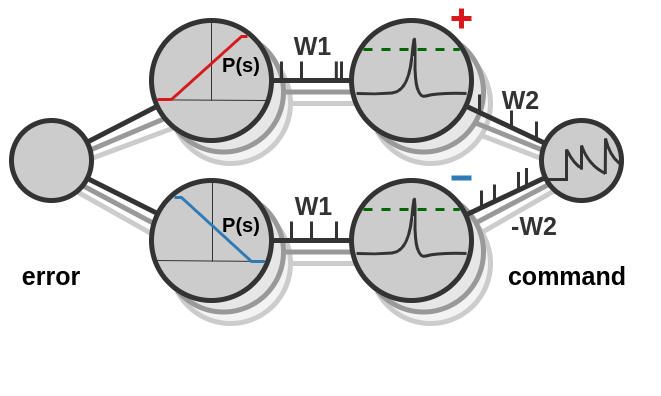}
    \caption{Structure of the proportional neuron groups. The value is encoded into "positive" and "negative" spikes and sent via symmetrical weights to a layer of Leaky Integrate and Fire neurons. The spikes emitted by these neurons are sent to a Leaky Integrate output neuron, resulting in an exponential filtered output.}
    \label{fig:p_structure}
\end{figure}

% \begin{figure}[h]
%     \centering
%     \SetCoordinates[xAngle=-90,yAngle=0,zAngle=-180]
%     \begin{tikzpicture}[multilayer=3d]
%     \Vertex[x=1.0,layer=1]{In}
%     \Vertex[x=0.5,layer=2]{A}
%     \Vertex[x=1.5,layer=2]{B}
%     \Vertex[x=0.5,layer=3]{C}
%     \Vertex[x=1.5,layer=3]{D}
%     \Vertex[x=1.0,layer=4]{Out}
%     \Edge(In)(A)
%     \Edge(In)(B)
%     \Edge(A)(C)
%     \Edge(B)(D)
%     \Edge(C)(Out)
%     \Edge(D)(Out)
%     \end{tikzpicture}
%     \caption{Structure of a single proportional neuron group}
%     \label{fig:p_structure}
% \end{figure}

\subsection{Integral - ensuring zero steady-state error}
The integral neurons should remove the steady-state error, unresolved by the proportional control. 
Initially, one may think that LIF neurons could integrate by setting the decay parameter to one (no decay). 
However, as soon as a spike occurs the integrated membrane potential is reset. 
Hence, if the membrane potential is not read out directly and the integrator has to be encoded by spikes, a different mechanism is required. 
% A simple feedforward LIF neuron has no way to integrate errors over time, thus a different solution is proposed in this work inspired by threshold adaptation mechanisms.
We propose a different solution in this work inspired by threshold adaptation mechanisms and the modulatory effects certain receptors exhibit.
There is a synaptic connection between the positive and negative encoding neurons to both neurons, as opposed to the proportional connections where there is only a connection between the positive neurons on the one hand and between the negative neurons on the other.
Besides the synaptic weights, there is also a signal flowing from the encoding neurons to the thresholds of the neurons in the integration layer, increasing or decreasing the threshold based on the sign. 
Previous work has studied various mechanisms for adapting the threshold based on inputs, often to prevent spike saturation. 
For example, the regular Adaptive LIF (ALIF), adapts its threshold based on its own spike activity~\cite{brette2005adaptive}. 
As a variation of the ALIF, 
%the authors of~
\citet{paredes2019unsupervised} proposes to deduct the pre-synaptic spike trace from incoming spikes, to discern features under varying input statistics, such as the per-pixel firing rate of event camera data.
In our method, however, the threshold is regulated based on weighing the incoming activity. 
A spike in the \textit{positive} integration neuron coming from the \textit{positive} encoding neuron decreases the threshold with a weight of $\theta_\text{add}$ (therefore increasing the spiking rate), while the \textit{negative} encoding neuron causes an increase with $\theta_\text{add}$ (thus decreasing the spiking rate), and vice-versa for the \textit{negative} integration neuron.  
This results in the following update rule for the threshold:
\begin{equation}
    \theta^\text{thr}(t+1) = \theta^\text{thr}(t) + \theta_{\text{add}} (s_{-}(t) -  s_{+}(t))
\end{equation}
with $s_{-}(t)$ and $s_{+}(t)$ the negative and positive incoming spikes, respectively.
Now, the encoding neurons act as a constant driving synaptic signal to maintain a certain activity in the integration layer, while the actual spiking rate is governed by the variation in the threshold. 
A common problem with PID regulators is integral windup, where actuator saturation or large changes in setpoint might lead to large amounts of accumulated error \cite{astrom1989integrator}. 
In our LIF model, we solve this by limiting the amount of change in the threshold, keeping the threshold in the range of $[0, 2\theta^\text{thr}]$ where $\theta^\text{thr}$ is the base threshold.
If the threshold is zero, the integrator will spike with the maximal spiking frequency, which is determined by the time step and refractory period.
\begin{figure}[h]
    \centering
    \includegraphics[width=0.8\linewidth]{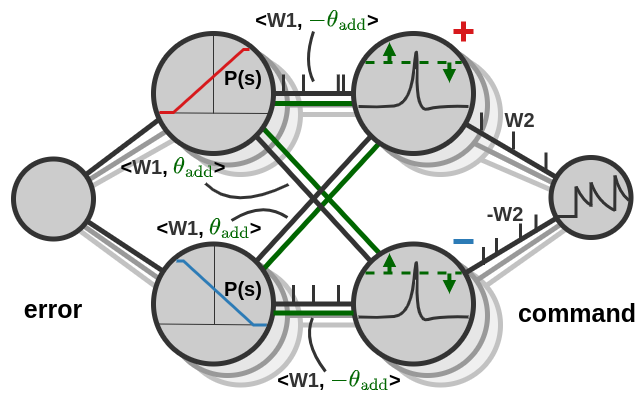}
    \caption{Structure of the integral neuron groups, bearing the Input-Weighted Threshold Adaptation (IWTA) mechanism. Next to the regular synapses, additional connections have been added that adapt the threshold of the LIF neurons according to their weights.}
    \label{fig:i_structure}
\end{figure}
It could be noted that multiple changes to this model are imaginable.
If a decay term is added to the threshold, it more closely resembles the ALIF, where the threshold converges back to a base value. % where decay / a decay term (not a decay)
In biology, this kind of input adaptation is similar to certain neurotransmitters with modulatory effects~\cite{di1994modulatory}.
Also, every input group now uses only a single positive and negative encoding neuron, with one update parameter $\theta_\text{add}$.
One could imagine using a larger group of encoding neurons, separate update weights $\theta_\text{add}$ per input connection and including these weights in the training procedure.
In this work, this remains unexplored since it focused on the task of integrating errors for which the proposed set of connections was sufficient. 

% \begin{figure}[h]
%     \centering
%     \SetCoordinates[xAngle=-90,yAngle=0,zAngle=-180]
%     \begin{tikzpicture}[multilayer=3d]
%     \Vertex[x=1.0,IdAsLabel,layer=1]{In}
%     \Vertex[x=0.5,IdAsLabel,layer=2]{A}
%     \Vertex[x=1.5,IdAsLabel,layer=2]{B}
%     \Vertex[x=0.5,IdAsLabel,layer=3]{C}
%     \Vertex[x=1.5,IdAsLabel,layer=3]{D}
%     \Vertex[x=1.0,IdAsLabel,layer=4]{Out}
%     \Edge(In)(A)
%     \Edge(In)(B)
%     \Edge[label=$wi_i$](A)(C)
%     \Edge[label=$wi_i$,distance=0.8](A)(D)
%     \Edge[label=$wi_i$](B)(D)
%     \Edge[label=$wi_i$,distance=0.8](B)(C)
%     \Edge[label=$wo_i$](C)(Out)
%     \Edge[label=$-wo_i$](D)(Out)
%     \end{tikzpicture}
%     \caption{Structure of a single integral neuron group}
%     \label{fig:i_structure}
% \end{figure}

\subsection{Derivative - decreasing overshoot}
The derivative action should be proportional to the change over time, countering any potential overshoot.
To obtain this, a similar structure to the proportional groups is used (as in Figure~\ref{fig:p_structure}), but now two of these groups are used in unison, instead of one. 
One of these groups has very fast time constants (higher weights, but faster decay), allowing it to react quickly proportionally to the input. 
The other has slower time constants (lower weights, but slow decay), resulting in an output that is a slightly delayed version of the input. 
By taking the difference between these two groups, we get a measure of the change over time which can be used as the derivative term in our PID controller.
Using multiple of these derivative groups again increases the accuracy of the overall estimate. For the derivative, this is especially important, as the derivative is usually already quite noisy due to noisy sensor measurements.

% \begin{figure}[h]
%     \centering
%     \SetCoordinates[xAngle=-90,yAngle=0,zAngle=-180]
%     \begin{tikzpicture}[multilayer=3d]
%     \Vertex[x=1.15,IdAsLabel,layer=1]{In}
%     \Vertex[x=0.5,IdAsLabel,layer=2]{A1}
%     \Vertex[x=0.8,IdAsLabel,layer=2]{A2}
%     \Vertex[x=1.5,IdAsLabel,layer=2]{B1}
%     \Vertex[x=1.8,IdAsLabel,layer=2]{B2}
%     \Vertex[x=0.5,IdAsLabel,layer=3]{C1}
%     \Vertex[x=0.8,IdAsLabel,layer=3]{C2}
%     \Vertex[x=1.5,IdAsLabel,layer=3]{D1}
%     \Vertex[x=1.8,IdAsLabel,layer=3]{D2}
%     \Vertex[x=1.15,IdAsLabel,layer=4]{Out}
%     \Edge(In)(A1)
%     \Edge(In)(A2)
%     \Edge(In)(B1)
%     \Edge(In)(B2)
%     \Edge[label=$wi_i$](A1)(C1)
%     \Edge[label=$wi_i$](A2)(C2)
%     \Edge[label=$wi_i$](B1)(D1)
%     \Edge[label=$wi_i$](B2)(D2)
%     \Edge[label=$wo_i$,distance=0.3](C1)(Out)
%     \Edge[label=$-wo_i$,distance=0.7](C2)(Out)
%     \Edge[label=$-wo_i$,distance=0.7](D1)(Out)
%     \Edge[label=$wo_i$,distance=0.3](D2)(Out)
%     \end{tikzpicture}
%     \caption{Structure of a single derivative neuron group}
%     \label{fig:d_structure}
% \end{figure}

\subsection{Training and tuning of the network}
Since the network has a substantial number of parameters that all influence the performance of the controller (such as synaptic weights, decay parameters and encoding parameters) and the parameters all depend on the time constants and gains of a specific controller, manual tuning is undesirable.
Therefore, it was chosen to use the architecture of the network as described above as a starting point and fine-tune the parameters using Backpropagation-Through-Time (BPTT). 
Because the LIF threshold function is non-differentiable, it was chosen to apply a surrogate gradient in the backwards-pass of BPTT~\cite{neftci2019surrogate}. Specifically, the surrogate gradient used in this work is the derivative of the arc-tangent as was proposed in~\cite{fang2021incorporating}. 

To force a response that is close to that of the target, the Mean Squared Error (MSE) was used as the dominant term in the cost function. 
Since during training, especially the derivative term was very sensitive to converging to local minima, it was chosen to add a second cost term for the derivative based on the Pearson correlation coefficient $\rho(x, \hat{x})$~\cite{cohen2009pearson}. 
Since we have a minimization problem and the perfect coefficient $\rho$ is 1, we use $1 - \rho(x, \hat{x})$ as the cost, further referring to it as the Pearson loss. 
For derivative control, it is very important that the control action is at least of the correct \textit{sign} because failing so might mean instabilities can arise. 
The Pearson loss promotes a linear relationship between both inputs and therefore supports the network to produce the correct sign. 
Finally, the parameters of the network need to stay within certain bounds, decay parameters for instance cannot be larger than 1 or smaller than 0. To force parameters to stay within these bounds, a linear exterior penalty function $\text{p}_\text{err}(\phi)$ is added to the cost function equal to the distance to the boundary. 
This results in the following cost function used in the BPTT training algorithm:
\begin{equation}
    J(\phi) = \text{MSE}(x, \hat{x}) + (1 - \rho(x, \hat{x})) + \text{p}_\text{err}(\phi)
\end{equation}
, where $x$ and $\hat{x}$ are the measured- and estimated values respectively, $1 - \rho(x, \hat{x})$ the Pearson loss. $\text{p}_\text{err}(\phi)$ is the error based on the parameters $\phi$, which is zero for all values of $\phi$ inside their specific range but of size $|\phi|$ for those outside. A table of all parameters included in the training, together with their valid ranges, can be found in the supplementary information. 

The data used for training was accumulated by logging the PID values of the regular Crazyflie controller on an onboard $\mu$SD card during manual flight. 
Care was taken to excite the system enough to gather a broad range of possible values a controller might encounter.

\section{Analysis SNN controller}
First, we look at the suitability of the multiple time constants for differentiation and afterward, we evaluate the IWTA mechanism for integral control.
\subsection{Derivative}
\label{sec:derivative}
\begin{figure}[h!]
    \centering
    \includegraphics[width=0.8\linewidth]{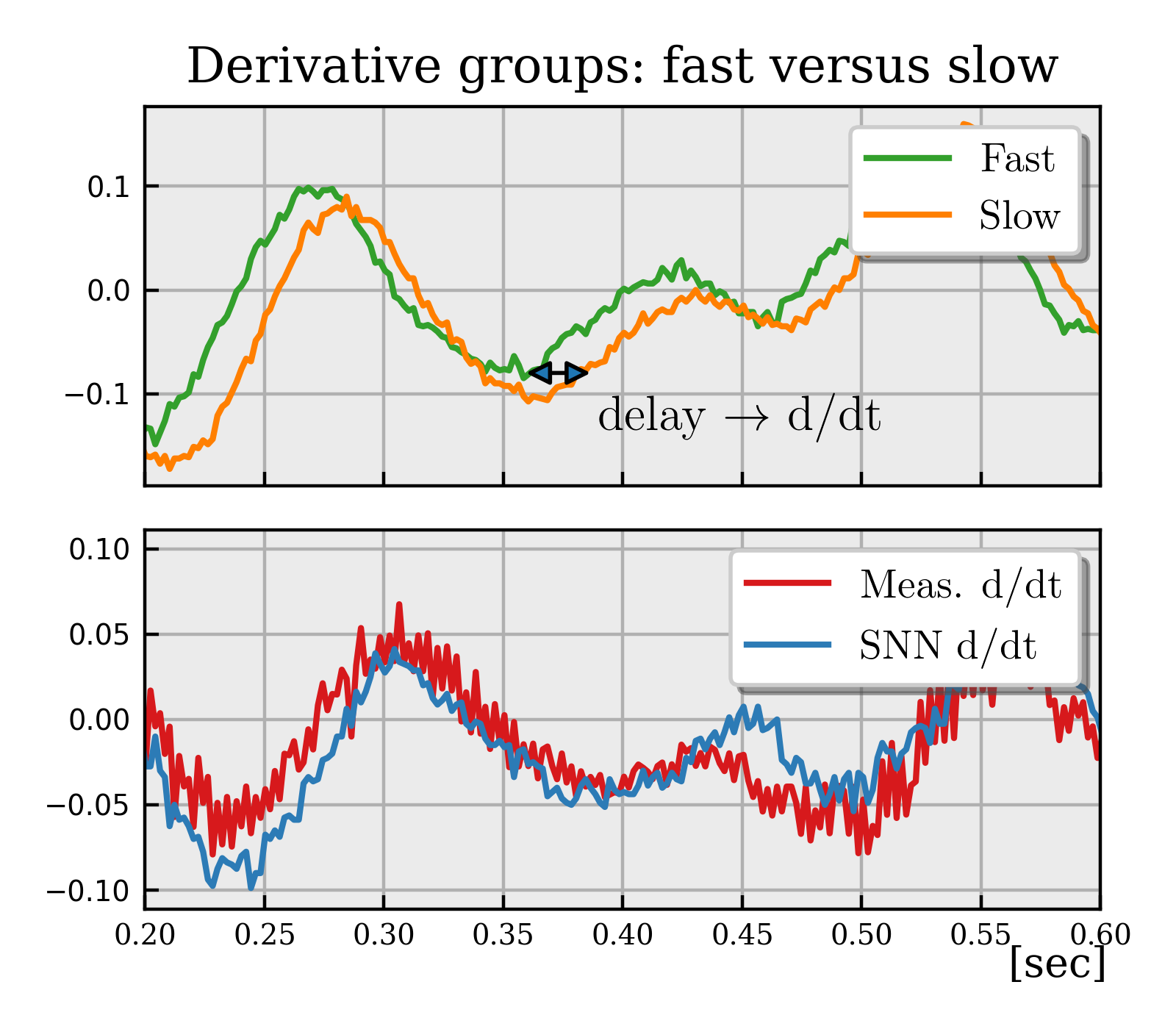}
    \caption{Example response to a common gyroscope input sequence recorded with the Crazyflie of both derivative groups. The top graph shows the average spiking rate of the fast and slow groups. As can be seen, the slow group is slightly delayed and the difference between both can be seen as a measure of the rate of change over time. In the bottom figure, this is visible as the derivative estimated with the SNN is compared to the measured derivative. }
    \label{fig:derivative}
\end{figure}

\begin{figure}[t]
    \centering
    \includegraphics[width=\linewidth]{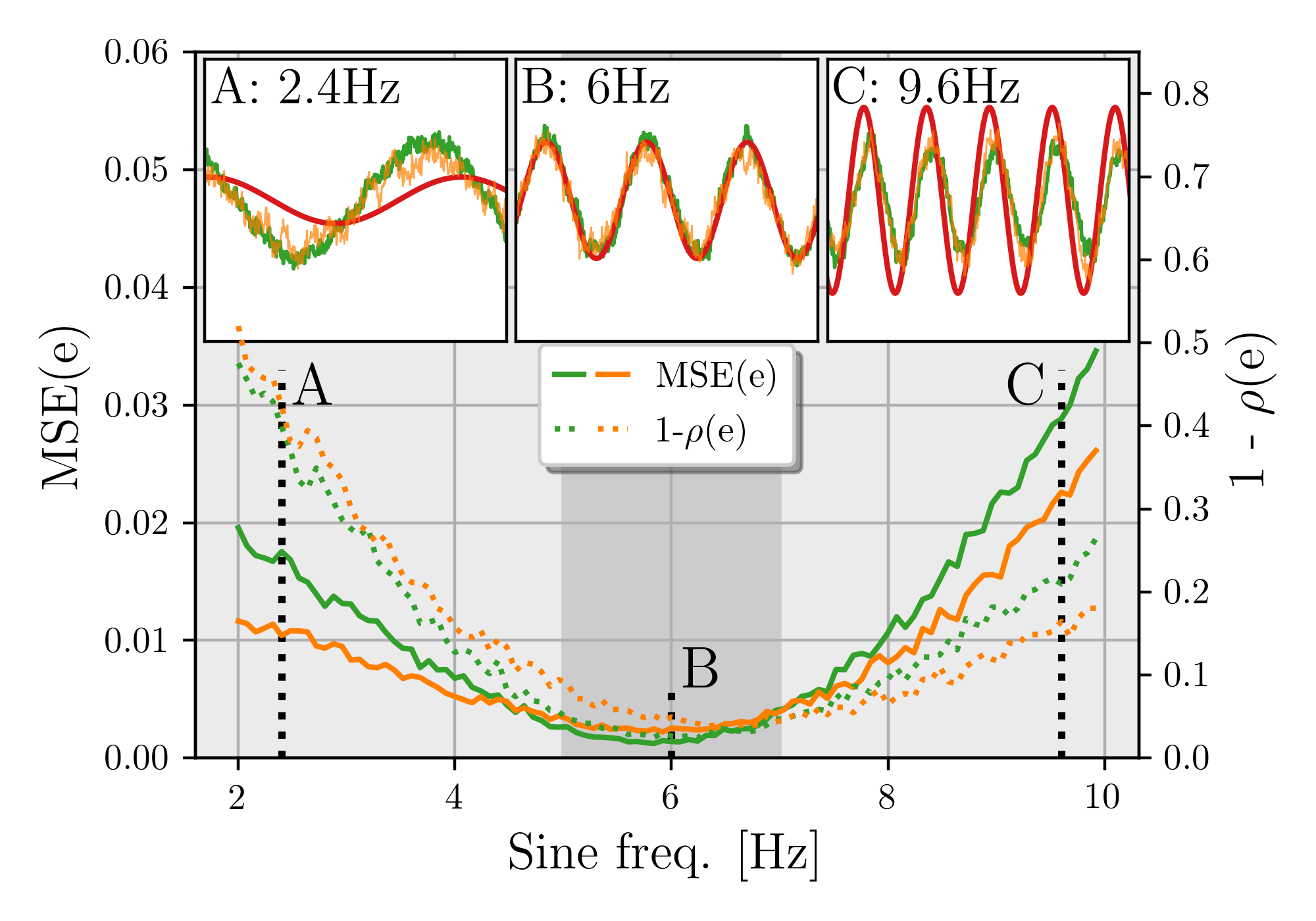}
    \caption{MSE and Pearson loss for a range of derivatives of sine waves after training on two sets of frequencies, 5-7Hz and 2-10Hz. Green lines correspond to the smaller range, and orange to the larger while the solid lines show MSE and dotted Pearson loss. The darker-colored area depicts the range on which the smaller set was trained. Three sub-figures are also included, demonstrating the response to a frequency A) below, B) inside and C) above the smaller training range with the SNN response in green and orange and real derivative in red. }
    \label{fig:derivative_sine_loss}
\end{figure}

To assess the ability of a network of LIFs with different time constants to estimate the derivative we start by looking at the response of both the fast and slow groups to an illustrational gyroscope sequence, obtained with the Crazyflie, after training. 
As can be seen in Figure~\ref{fig:derivative}, the average spiking rate of the slow groups is approximately a delayed version of the fast groups. 
By subtracting the delayed version we obtain a result similar to first-order backward difference, usually used in robotics to calculate the derivative of sensor data.
We noticed, however, that a particular set of time constants fits chiefly to the data it was trained upon. 
To further investigate this behavior, the response to sine waves with different frequencies was studied. 
Figure~\ref{fig:derivative_sine_loss} shows the MSE and Pearson loss for a range of sine waves with different frequencies after training on two sets of frequencies. 
The network was trained on a smaller range of sine waves between 5 and 7Hz, and also on a much wider range of 2 to 10Hz. 
Afterwards, the response to the entire frequency range was compared for both trained networks.
Although the network can accurately determine the derivative for the middle frequencies (B), it is too fast for lower frequencies (A), where the network in blue rises to its peak faster than the real derivative in red, and too slow for higher ones (C) where the network reaches its tipping point later. 
It is also visible that even with the larger range of input frequencies during training, the different time constants cannot correctly represent the entire band. 
Although the overall error gets reduced when training on the large band, the response for the lower and higher frequencies is still inaccurate.
This suggests that a different mechanism would need to be introduced in order to obtain perfect differentiation independent of input frequency. 

The importance of both cost terms in the loss function is also visible; the MSE cost for lower frequencies is relatively lower than the Pearson loss and for higher frequencies vice versa. 
Also, the amplitude of the SNN does not scale with the frequency, as is the case in the real derivative. 
This asserts the need for a reliable training procedure to tune the network to the application used, as was done in this work. 

\subsection{Integral}
\begin{figure}[h!]
    \centering
    \includegraphics[width=\linewidth]{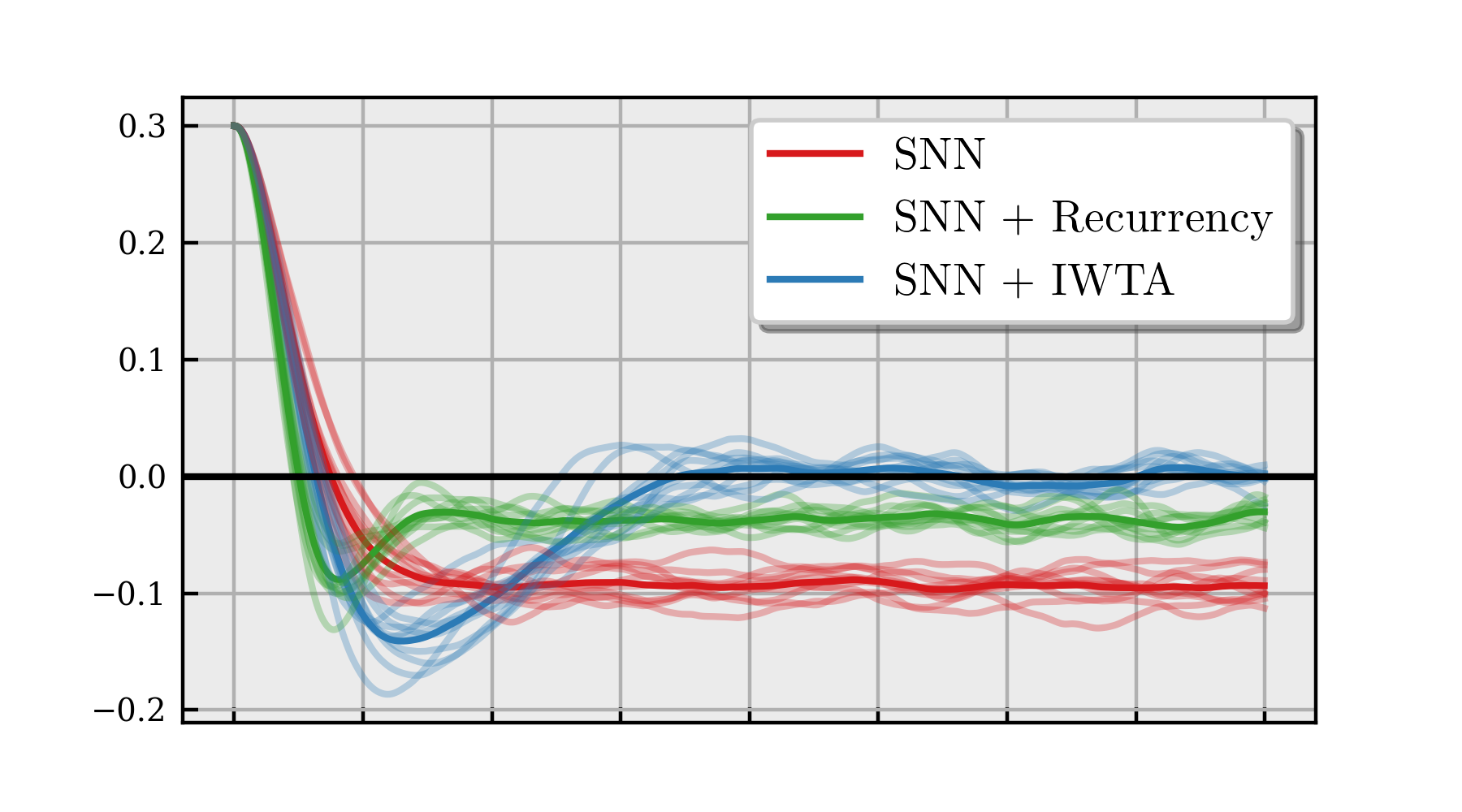}
    \caption{Comparison of three controllers to a setpoint-control task with an unknown input disturbance. The SNN with PD control only shows a constant offset from the setpoint after converging. The SNN with recurrency is clearly able to reduce the steady-state error, but due to representation error retains a small offset. The network equipped with Input-Weighted Threshold Adaptation effectively removes the steady-state error.}
    \label{fig:integrator_comparison}
\end{figure}

Furthermore, we have looked into the potential of the proposed Input-Weighted Threshold Adaptation (IWTA) mechanism. 
To present the importance of a functioning integrator, the classical control problem of the double integrator is used as an example. 
We have implemented the discrete-time equations of the double-integrator as:
\begin{align}
    x(t + dt) &= 
    \begin{bmatrix}
        1 & dt\\
        0 & 1
    \end{bmatrix}
    x(t) + \begin{bmatrix}
        \frac{1}{2} dt^2 \\
        dt
    \end{bmatrix}
    (u(t) - g) \\
    y(t) &= \begin{bmatrix}
        1 & 0
    \end{bmatrix} x(t)
\end{align}
, where $u(t)$ is the control input and $g$ is a constant input disturbance. 
In Figure~\ref{fig:integrator_comparison} we show the response of this system to three different SNNs; the LIF SNN with only PD components, an SNN with PD components and an integrator group that has fully recurrent connections (as was used in~\cite{zaidel2021neuropid}) and lastly our SNN with IWTA. 
The system starts from an initial state of $x0=[0.3, 0.0]^\text{T}$ and gets a setpoint $0.0$. The constant disturbance force $g$ was chosen to be $4.0$.
First, the SNN without any integrating mechanism settles at a steady-state error of $-0.1$. 
Second, the SNN with recurrent connections manages to reduce this steady-state offset, but not remove it completely. Due to representation errors, the recurrent connections cannot perfectly describe the integrated error. This means that the accumulated error is either underrepresented or overrepresented. In the former case, the integrator leaks information each time step and thus never completely removes the steady-state error. The green lines in Figure~\ref{fig:integrator_comparison} show this, since it clearly reduces the steady-state error, but does not remove it completely.
In the latter, the feedback amplifies the error at each time step which makes the system unstable and leads to all integrator neurons spiking at each step. 
Finally, the network that uses IWTA reaches a zero steady-state error. 

\section{Real-world experiments}
After the components are evaluated separately in simulation on toy problems, the use of the complete network was verified on a real-world problem.
\subsection{Hardware implementation}
To demonstrate the capabilities of our approach, we have implemented it as the lowest control layer of the tiny open-source quadrotor Crazyflie~\cite{giernacki2017crazyflie} (See Figure~\ref{fig:crazyflie}). 
The Crazyflie was enriched with enough computational power by the development of a deck module based on a Teensy 4.0 development board allowing the SNN to run in real-time (500Hz) in C++ on the ARM Cortex-M7 microprocessor. 
For the real-world tests, two scenarios were discerned; 1) manual flight and 2) position control. 
For manual control the Crazyflie receives attitude (roll/pitch/yaw) commands from a (manually controlled) radio transmitter and the higher-level attitude controller transforms these into rate setpoints. 
In the case of position control, the Crazyflie receives position measurements obtained with a Motion Capture system along with position commands via radio and via the same high-level controller these are converted to rate setpoints. 
These setpoints are sent, along with the gyroscope measurements from the onboard Inertial Measurement Unit (IMU), via UART to the deck where the SNN controller is evaluated at 500Hz. 
The torque command outputs of the neural controller are in turn sent back to the Crazyflie via the same UART connection, where they are inserted directly in the motor mixer. 
The total take-off weight of the Crazyflie including the Teensy 4.0 is only 35 grams, allowing for approximately 5 minutes of flight time.

% \begin{figure}
%     \centering
%     \includegraphics[width=\linewidth]{images/snnintegrator.png}
%     \caption{Overview of the hardware architecture that was used to test the proposed controller. Roll, pitch, yaw commands arrive from the connected transmitter on the Crazyflie and are combined within the attitude controller into a desired roll-, pitch- and yawrates, which are sent to the Teensy 4.0 mounted on the bottom of the Crazyflie via a serial connection together with the measured rates from the onboard Inertial Measurement Unit (IMU). Combined, they are fed into the neural controller, which outputs the desired motor setpoints. These are immediately sent back via the same serial connection to the motor mixer on the Crazyflie to produce motor commands at the desired update rate of 500Hz.}
%     \label{fig:hardware_architecture}
% \end{figure}

\begin{figure}[ht]
    \centering
    \includegraphics[width=\linewidth]{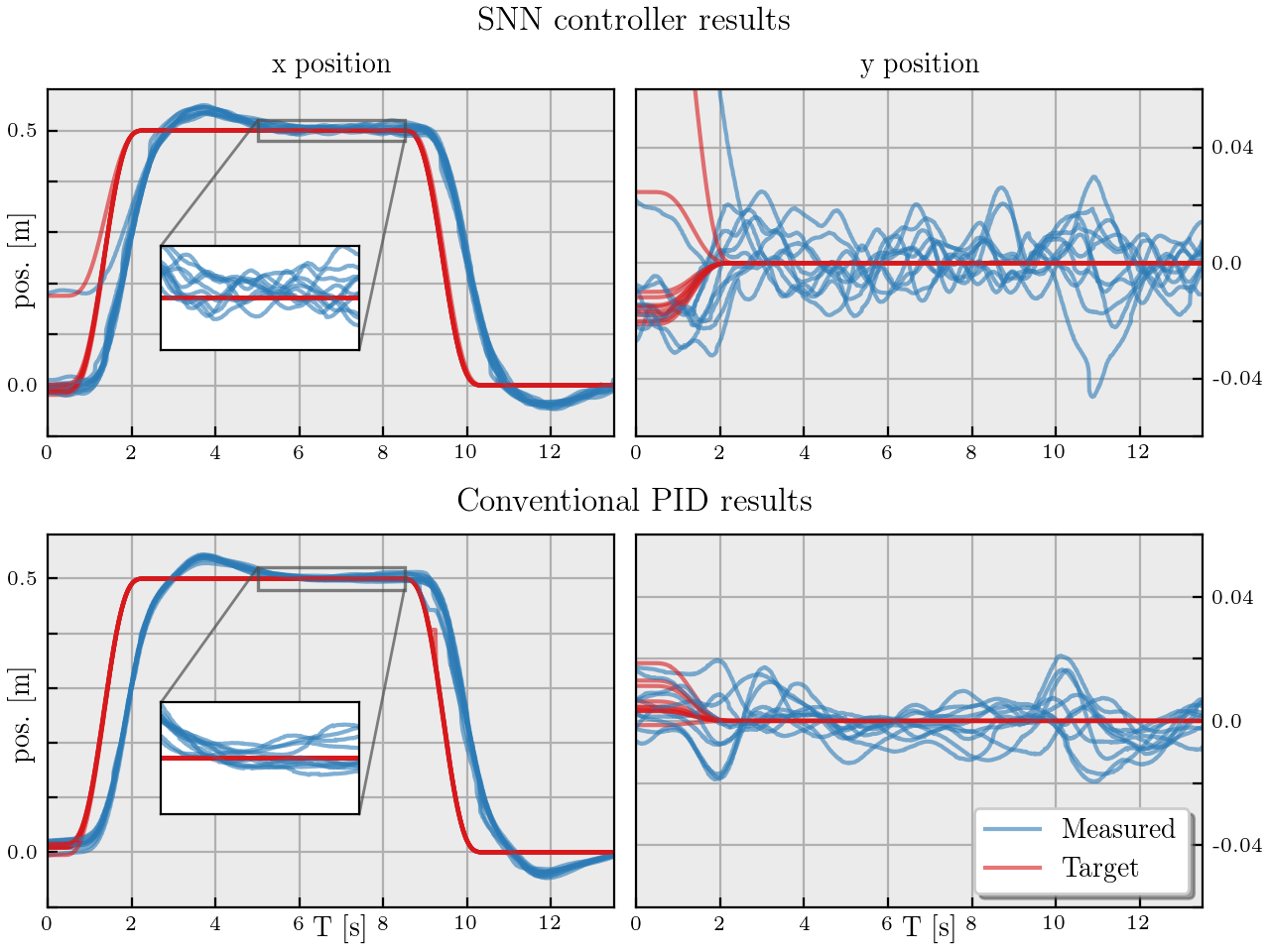}
    \caption{Position setpoint responses of both the SNN controller (top) and the conventional PID (bottom). For both, the positions as measured by the Motion Capture system are shown in blue, and setpoints in red. Although the SNN controller has a slightly noisier response, its trajectory is very similar to the PID.}
    \label{fig:pid_vs_snn}
\end{figure}

\subsection{Flight tests}
\begin{figure}[ht]
    \centering
    \includegraphics[width=0.8\linewidth]{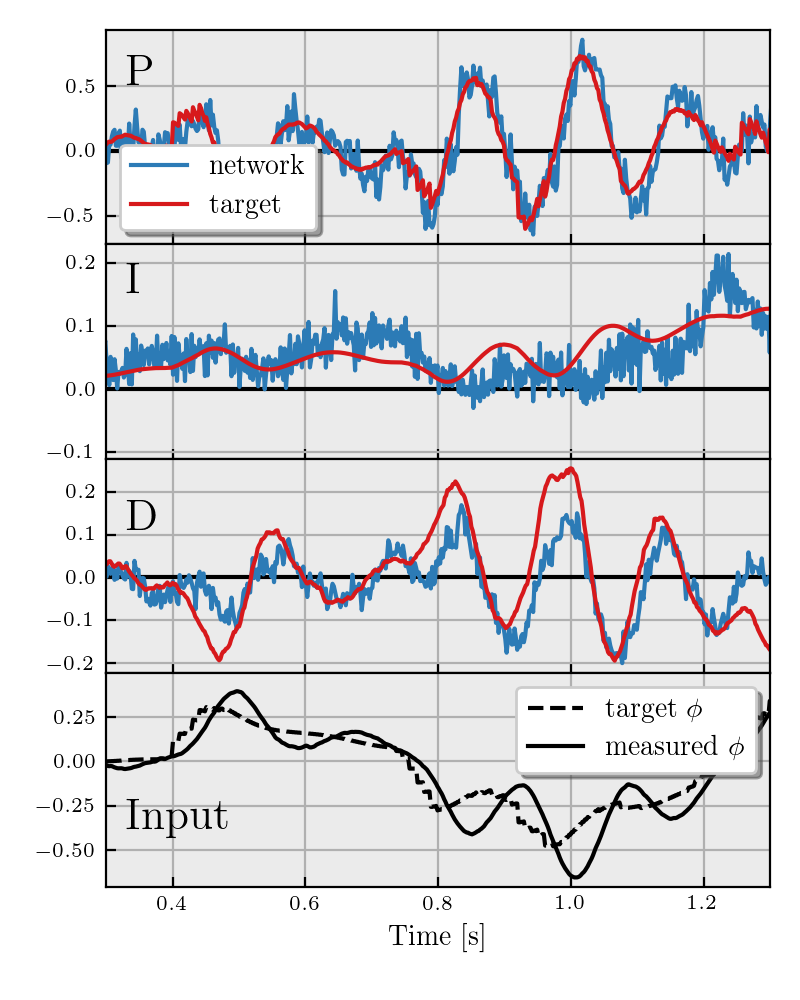}
    \caption{Output of the SNN controller in blue versus target values computed with a conventional PID controller in red for a manually controlled flight. The three uppermost figures depict the individual components of the controller. As can be seen, the SNN controller produces very similar results to the reference controller. The lowermost figure shows the input to the network; the target rotational rate $\phi$ and the rate measured by the gyro.}
    \label{fig:output_comparison}
\end{figure}

To demonstrate the capabilities of the network, a position-control test was performed along with manual flight. 
First, the Crazyflie was ordered to take off at its current position and after 2 seconds it was ordered to move to [0.5, 0.0] in 2 seconds.
This test was repeated 10 times for both the SNN controller and the regular PID controller as a benchmark. 
The results of all these 10 tests are shown in Figure~\ref{fig:pid_vs_snn}.
In the top two plots, the response of the SNN controller is shown. 
Among all ten tests, the controller remained stable and was able to follow the setpoint. 
The trajectory followed is very similar to that of the regular PID controller. 
However, on the $y$-results, as well as the inset axes for $x$, it is visible that the SNN controller has slightly larger deviations around the setpoint.
These can be caused by the stochasticity in the encoder. 
Since the input floating point value is reduced to a binary spiking input, the accuracy of the encoding is influenced by the encoding parameters but mostly by the number of input groups used. 
For this work, only 40 groups were chosen per P, I and D pathway for each control axis. Since the P and I terms require 2 neurons per group and the D term 4, this sums up to a total of 320 neurons per controller. This small number of neurons was chosen to showcase the possibilities of using small-scale systems for neuromorphic control. 
Also, the response to manual flight was performed with which the response of the different parts of the controller on a real-world system could be analyzed. 
In Figure~\ref{fig:output_comparison}, one second of such a test is shown. 
It is evident that the proportional part is very accurate, and the integral pathway is able to effectively deal with prolonged errors. 
The response of the derivative pathway is less accurate. This is most likely caused by the mechanisms shown in the derivative analysis, earlier in this work (Section~\ref{sec:derivative}), where it was shown that the derivative pathway tunes to specific delays. 
Even though the network is trained on real flight data, the range in delays might be too large which results in larger errors for the lowest and highest frequencies.
For control of the Crazyflie, this is still acceptable since the derivative control path still effectively dampens the response by countering large derivatives in the input.

\section{Conclusion}
In this work, we have proposed a novel input threshold adaptation mechanism, Input-Weighted Threshold Adaptation (IWTA). This mechanism adds extra weights per input connection that regulate the spiking threshold of the LIF neuron. By doing so, it enhances the network with the ability to integrate information over time, something the regular LIF model is unable to do. 

Also, we have shown that neuromorphic controllers using rate-based encoding can be used to control highly unstable underactuated systems. 
To demonstrate this, we have shown control of the innermost loop of a real flying tiny quadrotor, the Crazyflie. 
Using only 320 neurons per control axis, the network showed to be capable of stable and robust control, with the potential of extremely low delays due to the high inference speed of neuromorphic hardware.
By a straightforward training method using surrogate gradients and Backpropagation Through-Time, the network can be fine-tuned to a very limited amount of data from a real-world flying drone. 
Due to the sparse connections, the network is able to optimally benefit from the advantages of neuromorphic hardware.

In future work, we intend to apply the IWTA mechanism to different tasks and benchmarks to further establish its potential. 
Even though the different time constants for the derivative neurons allow us to dampen the control response, we have shown that they are limited to a specific frequency. 
We will also investigate the application of IWTA on the derivative neurons to improve the accuracy over a much broader range of frequencies.
The availability of a neuromorphic controller, such as a PID, that can be easily implemented on neuromorphic hardware plays a crucial role in completing the neuromorphic control loop in robotics. These controllers can be readily integrated into pipelines utilizing event-based algorithms, like a vision-based control system using an event camera as in~\cite{paredes2023fully}.
Lastly, the fast-paced and unpredictable movements of drones demand high-performance computing that traditional hardware struggles to provide, making neuromorphic processing an attractive alternative.

% making it necessary to enhance conventional hardware to effectively support the development of neuromorphic computing in this domain.

%\addtolength{\textheight}{-10cm}   % This command serves to balance the column lengths

%%%%%%%%%%%%%%%%%%%%%%%%%%%%%%%%%%%%%%%%%%%%%%%%%%%%%%%%%%%%%%%%%%%%%%%%%%%%%%%%

% RESET COUNTERS FOR SUPPLEMENTARY INFORMATION
\beginsupplement

%%%%%%%%%%%%%%%%%%%%%%%%%%%%%%%%%%%%%%%%%%%%%%%%%%%%%%%%%%%%%%%%%%%%%%%%%%%%%%%%

%%%%%%%%%%%%%%%%%%%%%%%%%%%%%%%%%%%%%%%%%%%%%%%%%%%%%%%%%%%%%%%%%%%%%%%%%%%%%%%%
\section*{Supplementary information}

\subsection{Current-based Leaky-Integrate-and-Fire neuron}
\label{sm:cubalif}
The current-based Leaky-Integrate-and-Fire (CUBA-LIF) is widely used in literature, available in most SNN simulators and commonly used in neuromorphic hardware, such as Intel's Loihi~\cite{davies2018loihi}. The discrete-time dynamic equations of the LIF neuron are as follows:
\begin{align}
    \label{eq:LIF}
    \upsilon_i(t+1) &= \tau^{\text{mem}}_ i\upsilon_i(t) + i_i(t) \\
    i_i(t+1) &= \tau^{\text{syn}}_i i_i(t) + \sum w_{ij} s_j(t)
\end{align}
\noindent , where $\upsilon_i(t)$ is the membrane potential at time $t$, $\tau^{\text{mem}}_i \in [0,1]$ and $\tau^{\text{syn}}_i \in [0,1]$ the membrane and synaptic time constants, $i(t)$ the synaptic current at time $t$, $w_{ij}$ the synaptic weight between neurons $i$ and $j$, and $s_j$ a binary value representing either a spike or no spike coming from the pre-synaptic neuron $j$. To determine whether a neuron emits a spike, the membrane potential is reduced with the neurons firing threshold $\theta^{\text{thr}}_i$ and passed through the Heaviside step function to determine the output of the neuron:
\begin{equation}
    s_i(t) = H(\upsilon_i(t) - \theta^\text{thr}_i)=\begin{cases} 0, & \upsilon_i(t)  - \theta^\text{thr}_i \leq 0 \\ 1, & \upsilon_i(t)  - \theta^\text{thr}_i > 0 \end{cases}
    \label{eq:heaviside}
\end{equation}

When the Heaviside function resolves to 1 and the neuron emits a spike, the membrane potential $\upsilon_i(t)$ is reduced by the threshold value (in literature this is known as a soft reset~\cite{han2020rmp}). 

\subsection{Trained parameters and ranges}
In Table~\ref{tab:parameters}, all the parameters used in the network are given, along with the specified range and the number that was used in the Crazyflie application. 
\begin{table}[h]
    \centering
    \caption{All parameters trained in the network, their given range and how many are used for the real-world tests.}
    \begin{tabular}{@{}lllll@{}} 
    \toprule
    & Parameter & Range & Count \\ 
    \midrule
    \textbf{P} & $\tau_i$ (current decay) & [0, 1] & 80\\
    & $\tau_v$ (voltage decay) & [0, 1] & 80\\
    & $w_i$ (input weight) & [0, $\inf$] & 40\\
    & $w_o$ (output weight) & [0, $\inf$] & 40\\
    \textbf{I} & $\tau_i$ (current decay) & [0, 1] & 80\\
    & $\tau_v$ (voltage decay) & [0, 1] & 80\\
    & $w_i$ (input weight) & [0, $\inf$] & 40\\
    & $w_o$ (output weight) & [0, $\inf$] & 40\\
    \textbf{D} & $\tau_i$ (current decay) & [0, 1] & 160\\
    & $\tau_v$ (voltage decay) & [0, 1] & 160\\
    & $w_i$ (input weight) & [0, $\inf$] & 80\\
    & $w_o$ (output weight) & [0, $\inf$] & 80\\
    \bottomrule
    \end{tabular}
    \label{tab:parameters}
\end{table}

\begin{acks}
This material is based upon work supported by the Air Force Office of Scientific Research under award number FA8655-20-1-7044.
\end{acks}

\bibliographystyle{ACM-Reference-Format}
\bibliography{biblio}

\end{document}